\documentclass[10pt,twocolumn,letterpaper]{article}

\usepackage[pagenumbers]{cvpr} 

\definecolor{cvprblue}{rgb}{0.21,0.49,0.74}
\usepackage[pagebackref,breaklinks,colorlinks,allcolors=cvprblue]{hyperref}

\usepackage{booktabs}       

\usepackage{color,xcolor}
\usepackage{epsfig}
\usepackage{graphicx}

\usepackage{adjustbox}
\usepackage{array}
\usepackage{booktabs}
\usepackage{colortbl}
\usepackage{hhline}
\usepackage{multirow}
\usepackage{tabularx}
\usepackage{makecell}
\usepackage{float}

\usepackage{amsmath,amsfonts,amsthm,amssymb}
\usepackage{bm}
\usepackage{nicefrac}
\usepackage{microtype}

\usepackage{changepage}
\usepackage{extramarks}
\usepackage{fancyhdr}
\usepackage{lastpage}
\usepackage{soul}
\usepackage{xspace}

\usepackage{url}

\usepackage{algorithm, algorithmic}
\usepackage{enumerate}
\usepackage{enumitem}

\usepackage{etoolbox,siunitx}
\usepackage{afterpage}
\usepackage{subcaption}
\usepackage[title]{appendix}
\usepackage{overpic}
\usepackage{bbding}

\def\eg{\emph{e.g.}}

\def\paperID{16997} 
\def\confName{CVPR}
\def\confYear{2025}

\title{GCA-3D: Towards Generalized and Consistent Domain Adaptation of 3D Generators}

\author{Hengjia Li$^{1}$ \and Yang Liu$^{2}$ \and Yibo Zhao$^{1}$\and Haoran Cheng$^{1}$\and Yang Yang$^{1}$\and Linxuan Xia$^{1}$\and Zekai Luo$^{1}$\and Qibo Qiu$^{1}$ \and Boxi Wu$^{1}$ \and Tu Zheng$^{3}$ \and Zheng Yang$^{3}$\and Deng Cai$^{1}$ \\
{
\textsuperscript{1} {Zhejiang University} \;
\textsuperscript{2} {Alibaba Group} \;
\textsuperscript{3} {Fabu Inc.} \;}
}
\begin{document}

\twocolumn[{
\renewcommand\twocolumn[1][]{#1}
\maketitle
\begin{center}
    \centering
    \includegraphics[width=1\textwidth]{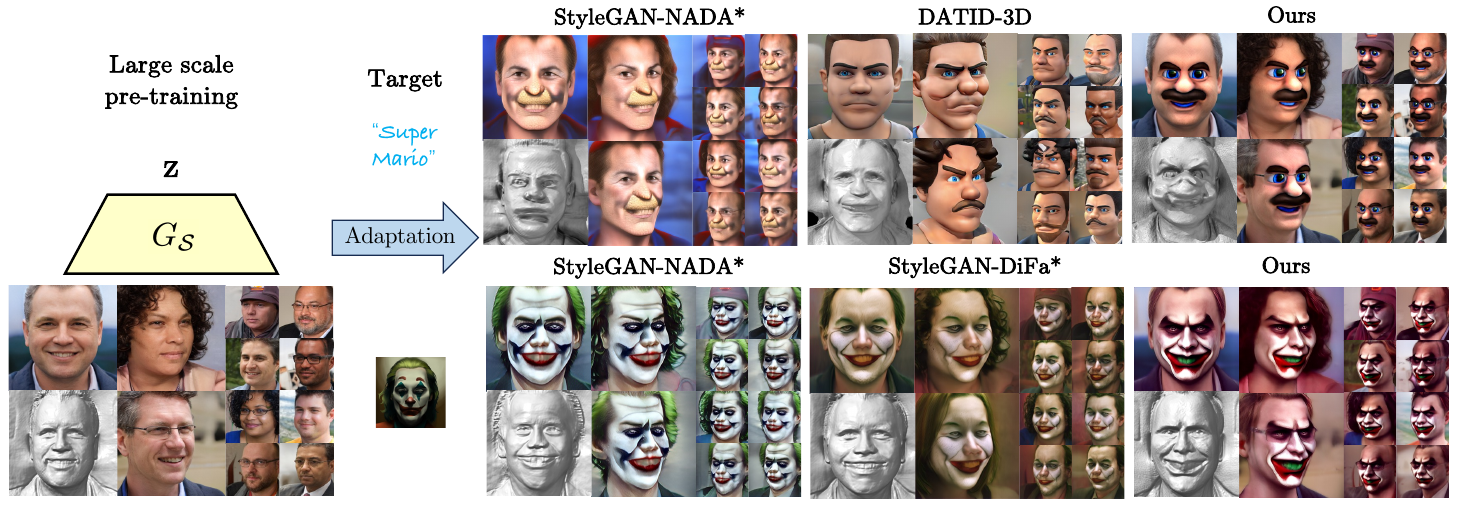}
    \captionof{figure}{
        GCA-3D is generalized to both text-guided and image-guided 3D generative domain adaptation. It successfully generate diverse results in the target domain, consistently maintaining \textit{accurate pose} and \textit{diverse identity} in the source domain, while the baseline methods fail. * means the 3D extension version of the 2D adaptation methods.
      }
\label{fig:visual}
\end{center}
\vspace*{.3cm}
}]

\begin{abstract}
Recently, 3D generative domain adaptation has emerged to adapt the pre-trained generator to other domains without collecting massive datasets and camera pose distributions. Typically, they leverage large-scale pre-trained text-to-image diffusion models to synthesize images for the target domain and then fine-tune the 3D model. However, they suffer from the tedious pipeline of data generation, which inevitably introduces \textit{pose bias} between the source domain and synthetic dataset. 
Furthermore, they are not generalized to support one-shot image-guided domain adaptation, which is more challenging due to the more severe pose bias and additional \textit{identity bias} introduced by the single image reference.
To address these issues, we propose GCA-3D, a generalized and consistent 3D domain adaptation method without the intricate pipeline of data generation.
Different from previous pipeline methods, we introduce multi-modal depth-aware score distillation sampling loss to efficiently adapt 3D generative models in a non-adversarial manner.
This multi-modal loss enables GCA-3D in both text prompt and one-shot image prompt adaptation.
Besides, it leverages per-instance depth maps from the volume rendering module to mitigate the overfitting problem and retain the diversity of results.
To enhance the pose and identity consistency, we further propose a hierarchical spatial consistency loss to align the spatial structure between the generated images in the source and target domain. 
Experiments demonstrate that GCA-3D outperforms previous methods in terms of efficiency, generalization, pose accuracy, and identity consistency. 
\end{abstract}    

\begin{figure}[ht]
\centering
\includegraphics[width=\linewidth]{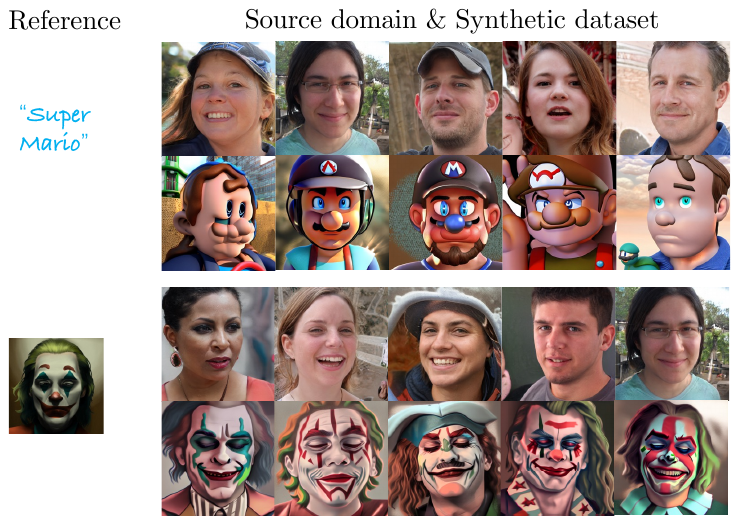}
\vspace{-0.4cm}
\caption{Bias issues in the synthetic dataset of pipeline methods like DATID-3D~\cite{kim2022datid} conditioned by single text prompt or one-shot image reference using the translation pipeline of Stable Diffusion~\cite{rombach2022high} with IP-Adapter~\cite{ye2023ip}. (1) In the data generation, pose bias introduced by the condition can affect pose consistency, particularly when the condition is a single image, leading to the generation of mostly \textit{forward-facing} images. (2) Attributes within the image reference, such as \textit{closed eyes in joker}, can also introduce identity bias, thereby impacting the diversity of the results.}
\label{fig:bias}
\vspace{-0.2cm}
\end{figure}

\section{Introduction}
3D generative models~\cite{chan2022efficient, liao2020towards, niemeyer2021giraffe} have been developed to facilitate the synthesis of multi-view consistent and pose-controlled images, which find extensive applications across various scenarios, including gaming, advertising, and digital human. 
Notably, some of them leverage the StyleGAN2~\cite{karras2020analyzing} generator in combination with neural rendering techniques, which achieve real-time generation of high-resolution, multi-view consistent images and detailed 3D shapes. 
However, training these 3D generative models necessitates not only a large dataset but also detailed information on camera pose distribution, which brings greater challenges than 2D models and restricts these models to limited domains.

Domain adaptation methods~\cite{gal2021stylegan, song2022diffusion, zhang2022towards, zhu2021mind, alanov2022hyperdomainnet, YotamNitzan2023DomainEO} like StyleGAN-NADA~\cite{gal2021stylegan} give a simple yet effective way to adapt the generator to diverse domains. They typically adopt CLIP~\cite{radford2021learning} or text-to-image diffusion models~\cite{rombach2022high} with non-adversarial fine-tuning as guidance of the adaptation. However, as shown in \cref{fig:visual}, they suffer from catastrophic loss of \textit{diversity}, which is inherent in a text prompt or one-shot image reference for non-adversarial fine-tuning~\cite{kim2022datid}. 

Recently, adversarial 3D generative domain adaptation methods~\cite{kim2022datid, kim2023podia} have emerged as promising solutions to alleviate the issue of diversity. They generally leverage text-to-image diffusion models with target prompts to generate and filter a large dataset to fine-tune the source domain's generator. While demonstrating diverse results, these methods suffer from the cumbersome pipeline of laborious data generation. These pipeline methods not only have a large computational cost, but also inevitably introduce \textit{pose bias} in the synthetic dataset and compromise the 3D pose accuracy. 

Furthermore, it becomes even more challenging for one-shot image-guided adaptation to generate images with the specific style or attributes of the image condition, which is under-explored but more flexible when the target domain is challenging to describe through text. As shown in \cref{fig:bias}, the pose of the image reference itself can exacerbate the pose bias between the generated dataset and the source domain. Additionally, it introduces \textit{identity bias} due to the reliance on a single image reference. For example, most faces in the synthesized dataset have closed eyes, which results in the adapted generator losing identity consistency as well as diversity.

To address these issues, we propose a generalized method called GCA-3D for pose and identity consistent 3D domain adaptation, which efficiently eliminates intricate pipeline of data generation. As shown in \cref{fig:visual}, GCA-3D consistently preserves the \textit{pose} and \textit{identity} with the source domain, which effectively improves the pose accuracy and diversity. In addition to text-driven adaptation, our GCA-3D is also the first trial to support one-shot image-guided adaptation, which addresses this gap in 3D generative domain adaptation. 

Specifically, drawing inspiration from the impressive performance of Score Distillation Sampling (SDS)~\cite{poole2022dreamfusion} in text-guided 3D generation, we introduce multi-modal depth-aware SDS loss (DSDS) for 3D generative domain adaptation.
To tackle the inherent overfitting problem in non-adversarial domain adaptation~\cite{kim2022datid}, we leverage the instance-aware depth map from the volume rendering module of the source generator as the condition during the adaptation. We also adopt the foreground mask from the depth map of the target generator to focus more on foreground adaptation and preserve the background information, which further enhances the diversity of generated images. Besides, to support image-guided domain adaptation, we further introduce IP-Adapter~\cite{ye2023ip} as the image encoder in DSDS loss, which achieves multi-modal 3D domain adaptation. 

To preserve the cross-domain consistency including pose and identity with the source domain, we propose a novel hierarchical spatial consistency loss (HSC) in GCA-3D to align the spatial structure with images from the source domain. Concretely, we leverage a patch-wise image encoder like MViTv2~\cite{fan2021multiscale} to encode the generated images into multi-scale patch tokens with hierarchical spatial structure information. To preserve the spatial consistency with the source domain, we align these patch tokens by adopting coarse-to-fine contrastive learning. As shown in \cref{fig:visual}, GCA-3D achieves superior 3D pose accuracy and identity consistency in domain adaptation. 

\begin{table}[tb]
    \centering
    \footnotesize
    \resizebox{\linewidth}{!}{
    \begin{tabular}{ccccc}
\toprule
Method & \makecell[c]{Training \\ Time} & \makecell[c]{Supportive \\Modality} & \makecell[c]{Pose \\Accuracy} & \makecell[c]{Identity \\Consistency} \\
\midrule
\emph{Adversarial methods} \\
\midrule

DATID-3D &9h & Text& \XSolidBrush & \XSolidBrush  \\
PODIA-3D$^\dag$ &- & Text& \Checkmark & \XSolidBrush    \\
\midrule
\emph{Non-adversarial methods} \\
\midrule
StyleGAN-NADA$^*$&1h &Image+Text &\XSolidBrush&\XSolidBrush 
 \\
StyleGAN-DiFa$^*$&1h &Image &\XSolidBrush&\Checkmark 
\\
StyleGAN-Fusion$^*$&1h &Text &\XSolidBrush&\XSolidBrush 
\\
DiffGAN3D$^\dag$&- &Text &\XSolidBrush&\XSolidBrush 
\\
GCA-3D&1h &Image+Text&\Checkmark&\Checkmark
 \\ 
\bottomrule
\end{tabular}}
\centering
\footnotesize
\caption{Comparison with previous methods. $*$ means the 3D extension version of the 2D adaptation methods and $\dag$ indicates that the method is not open source.}
\label{tab:sum}
\end{table}

We conduct experiments for a wide range of domains including both text-driven and image-driven, which demonstrates the effectiveness and generalization of our GCA-3D. To sum up, as shown in \cref{tab:sum}, GCA-3D significantly outperforms previous methods in terms of efficiency, generalization, pose accuracy, identity consistency, and diversity. Overall, our contributions are as follows:
\begin{itemize}
\item We propose an efficient framework for 3D generative domain adaptation, eliminating the bias of pose and identity and simplifying the tedious pipeline of data generation. We demonstrate its superiority over previous methods.
\item We introduce a multi-modal depth-aware SDS loss for 3D domain adaptation, which effectively overcomes the overfitting problem in non-adversarial adaptation. To the best of our knowledge, it is the first method explored for 3D image-guided domain adaptation.
\item We propose a novel hierarchical spatial consistency loss to preserve the consistency of pose and identity with the source domain. Extensive experiments show its effectiveness in both pose accuracy and diversity. 

\end{itemize}

\begin{figure*}[ht]
\centering
\includegraphics[width=.9\linewidth]{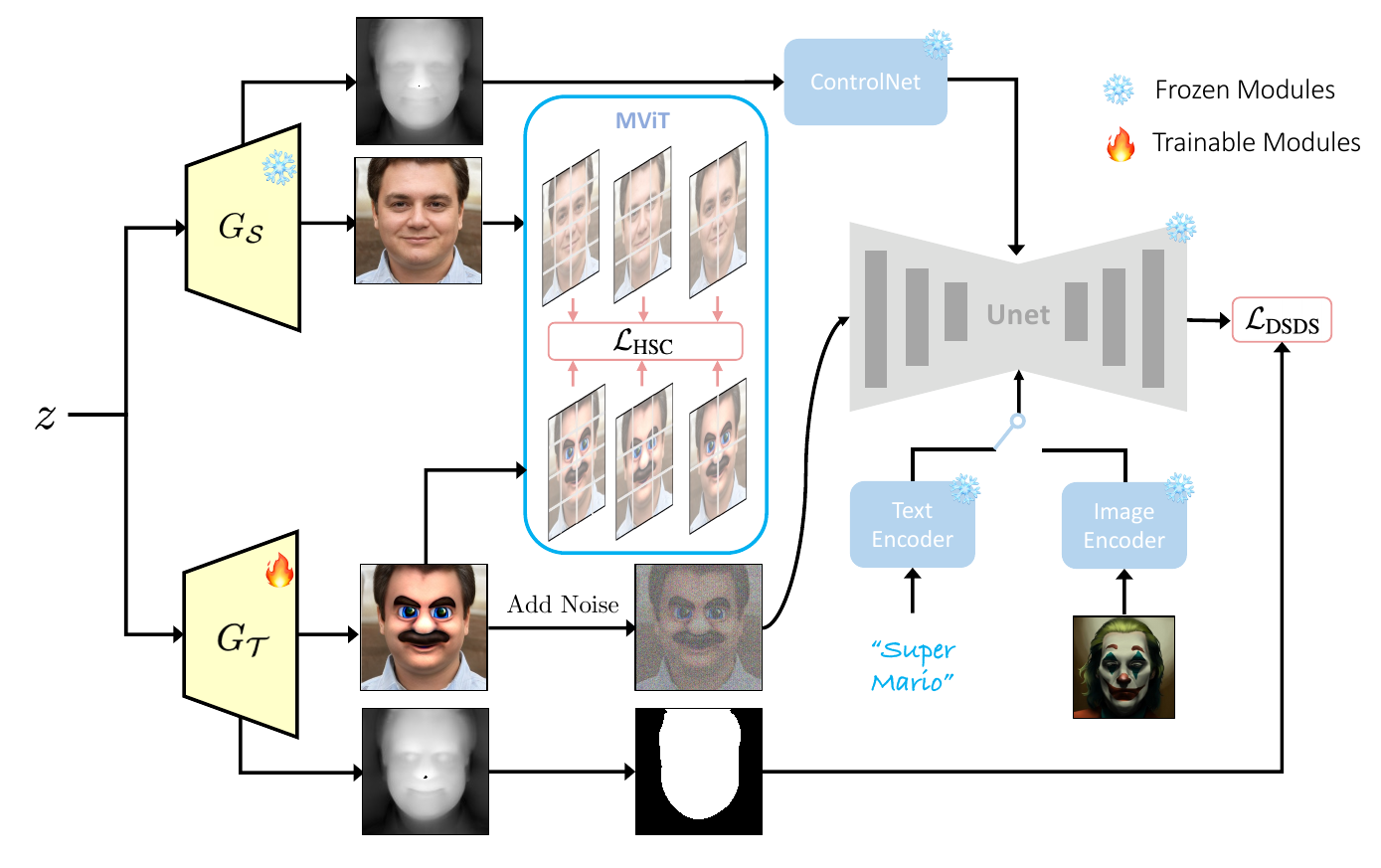}
\vspace{-0.4cm}
\caption{Overview of GCA-3D. Given the source generator 3D generator $G_\mathcal{S}$ and our target generator $G_\mathcal{T}$ (initialized from $G_\mathcal{S}$), we propose depth-aware SDS (DSDS) loss to enable multi-modal 3D domain adaptation via pre-trained CLIP encoder and IP-Adapter. To eliminate the pose and identity biases of the target domain, we propose Hierarchical Spatial Consistency (HSC) loss to coarse-to-fine align the synthesized images of $G_\mathcal{S}$ and $G_\mathcal{T}$ given the same noise $z$.}
\label{fig:method}
\vspace{-0.4cm}
\end{figure*}

\section{Related Work}
\subsection{2D Generative Domain Adaptation}
2D domain adaptation adapts image generators to target domains through text-driven or image-driven methods. Text-driven methods~\cite{alanov2022hyperdomainnet,gal2021stylegan,liu2023text,lyu2023deltaedit,YotamNitzan2023DomainEO,song2022diffusion,zhu2022one,li2024unihda} involves using a textual prompt to shift the domain of a pre-trained model toward a new domain.
For instance, StyleGAN-NADA~\cite{gal2021stylegan} refines pre-trained StyleGAN2~\cite{karras2020analyzing} by leveraging a straightforward textual prompt, guided by CLIP~\cite{radford2021learning}.
Image-driven generative domain adaptation~\cite{alanov2023styledomain,duan2024weditgan,kim2022diffface,li2020few,mo2020freeze,mondal2023few,ojha2021few,wu2024domain,xiao2022few,ye2023ip,zhang2022towards,zhao2022few,zhao2022closer,zhu2021mind,li2023few} involves adjusting a pre-trained image generator for a new target domain using a small set of training images. Some research use additional regularization to reduce overfitting. For example, DiFa~\cite{zhang2022towards} applies GAN inversion~\cite{tov2021designing} to align the latents, thereby ensuring diversity is maintained from the original generator.

\subsection{3D Generative Domain Adaptation}
3D generative models~\cite{liao2020towards, nguyen2020blockgan, chan2021pi, niemeyer2021giraffe, schwarz2020graf, gu2022stylenerf, zhou2021cips} have seen rapid development. Specially,
EG3D~\cite{chan2022efficient} uses a triplane representation, integrating StyleGAN2~\cite{karras2020analyzing} with neural rendering~\cite{mildenhall2020nerf} to generate high-quality 3D shapes and consistent images.
Recently, 3D generative domain adaptation~\cite{abdal20233davatargan,jin2022dr,zhang2023deformtoon3d,zhou2021cips} has been developed to adapt pretrained generators for other 3d domains without the need for extensive, meticulously annotated datasets.
Leveraging the powerful image generation capabilities of diffusion models~\cite{rombach2022high}, studies such as ~\cite{kim2022datid,kim2023podia,zhang2023styleavatar3d} have utilized diffusion models to generate training datasets in the target domain, achieving impressive results in text-guided 3D domain adaptation.
Additionally, non-adversarial fine-tuning methods ~\cite{alanov2022hyperdomainnet,gal2021stylegan,song2022diffusion,lei2024diffusiongan3d} have demonstrated significant potential in the realm of text-guided 3d domain adaptation.
Especially, StyleGAN-Fusion~\cite{song2022diffusion}, utilizes SDS loss\cite{poole2022dreamfusion} to guide the adaptation of both 2D and 3D generators. This method also faces challenges like limited 3D supervision and constraints with image input.

\subsection{Conditional Diffusion Models}
Diffusion models \cite{ho2020denoising,song2020score,nichol2021improved} bring a recent breakthrough in image generation. To produce images that align with specific requirements, recent research has integrated text and image into diffusion models.
For text conditioning, denoising is guided by textual embeddings generated by language encoders~\cite{radford2021learning,raffel2020exploring} and is performed either in pixel space~\cite{nichol2021glide,saharia2022photorealistic,ramesh2022hierarchical} or latent space~\cite{gu2022vector,rombach2022high}.
For image conditioning, 
some works~\cite{voynov2023sketch,zhang2023adding,mou2023t2i,mo2024freecontrol,xie2023boxdiff,yang2024lora,zhao2023local} incorporate image structural information, such as bounding boxes, canny edges, or depth, either through training \cite{voynov2023sketch,zhang2023adding,mou2023t2i,uni,li2024personalvideo} or without requiring additional training \cite{mo2024freecontrol,xie2023boxdiff}.
Another technique for using image prompts is applied in customized generation, where the goal is to extract information such as objects~\cite{ruiz2023dreambooth,xiao2023fastcomposer,ye2023ip} and styles~\cite{hertz2024style} from the image to regenerate a new one.
\section{Method}
\subsection{Preliminary}
\textbf{EG3D}~\citep{chan2022efficient}) is a state-of-the-art 3D generator to synthesis multi-view consistent and pose-controlled images. 
Specifically, it integrates StyleGAN2~\citep{karras2020analyzing} generator with the latent code and camera pose as input and a triplane decoder to synthesize triplane as 3D representation. Then it utilizes volume rendering~\citep{mildenhall2021nerf} with a super-resolution module to achieve high-quality 3D shapes and pose-controlled image synthesis. 

\noindent
\textbf{Score Sampling Distillation} (SDS)~\citep{poole2022dreamfusion} is an optimization method by distilling pretrained text-to-image diffusion models, which is widely used in text-to-3D generation. For an image $x$ generated by a 3D generator $G$ with parameter $\theta$, SDS adds random noise $\epsilon$ on $x$ at the timestep $t$ to obtain a noisy image $z_t$. Given a pre-trained text-to-image diffusion model with the noise prediction network $\epsilon_\phi$ and target text prompt $y$, SDS optimizes $\theta$ by:
\begin{equation} \label{eq:sds}
\nabla_{\theta} \mathcal{L}_{\text{SDS}}(\phi, G_{\theta}) = \mathbb{E}_{t, \epsilon} \left[ w_t \left( \epsilon_{\phi}(z_t; y, t) - \epsilon \right) \frac{\partial x}{\partial \theta} \right],
\end{equation}
where $w_t$ indicates a weighting function depending on $t$. Despite these advancements, empirical evidence~\citep{poole2022dreamfusion} indicates that SDS loss often suffers from low diversity, which has not yet been adequately addressed.

\subsection{Generalized 3D Domain Adaptation}
We start with a pre-trained 3D generator $G_\mathcal{S}$ (\eg, EG3D), that maps from noise $z$ with specific camera pose to the image $x$ in a source domain $\mathcal{S}$. Given a new target domain $\mathcal{T}$, generative domain adaptation aims to adapt $G_\mathcal{S}$ to yield a target generator $G_{\mathcal{T}}$, which can generate images similar to domain $\mathcal{T}$. Typically, we initialize $G_{\mathcal{T}}$ from $G_\mathcal{S}$ and fine-tune it during the adaptation.

Existing 3D adaptation methods~\cite{kim2022datid, kim2023podia, lei2024diffusiongan3d} are limited to text-guided adaptation, where the target domain is specified via text. However, this approach becomes less practical when the target domain is difficult to accurately describe using text alone, such as a specific artistic style. For more general purposes, we aim for generalized 3D domain adaptation in this paper, which enables both a single text prompt and a one-shot image reference.

\begin{figure*}[ht]
\centering
\includegraphics[width=\linewidth]{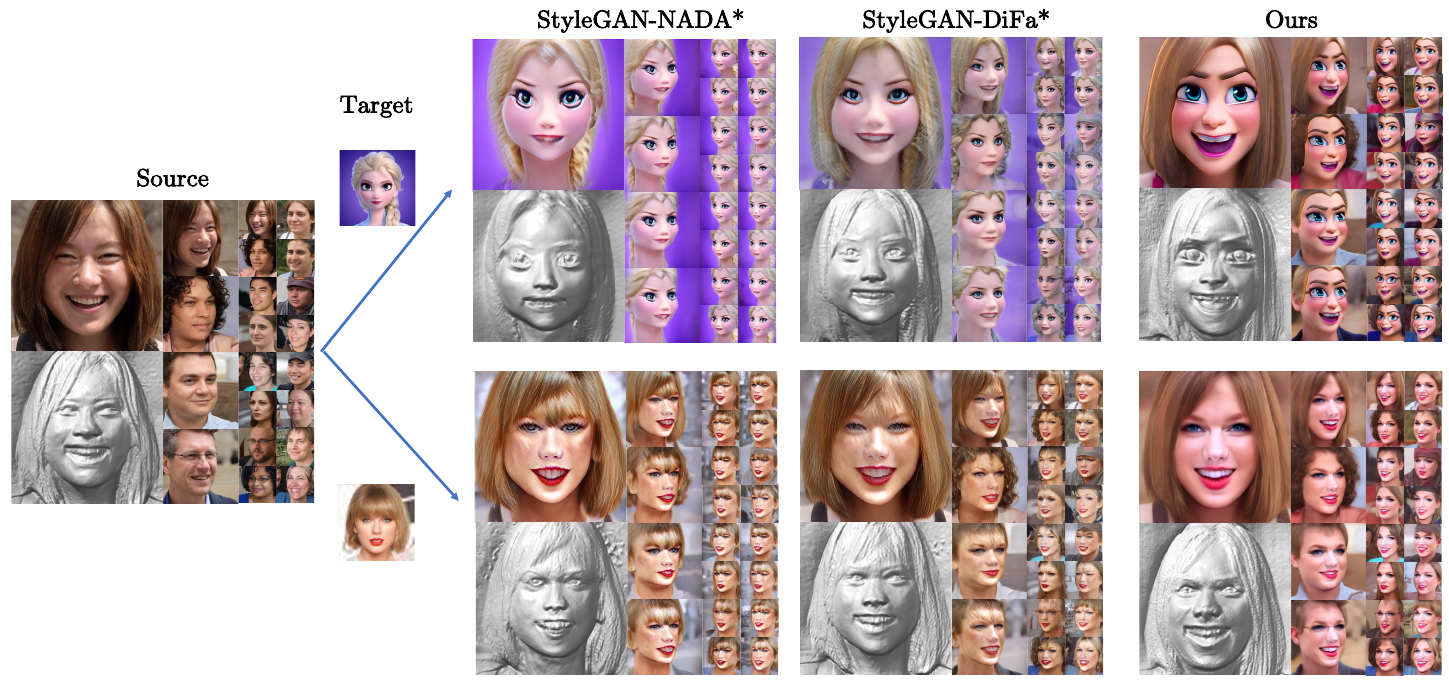}
\vspace{-0.6cm}
\caption{Qualitative comparison with existing image-guided domain adaptation methods. Our GCA-3D significantly surpasses baseline methods in diversity, pose accuracy and identity consistency.}
\label{fig:img}
\vspace{-0.4cm}
\end{figure*}

\subsection{Multi-Modal Depth-aware SDS loss}
Due to the challenges in acquiring high-quality pose-aware data, adversarial learning methods~\cite{kim2022datid, kim2023podia, lei2024diffusiongan3d} leverages diffusion models~\cite{rombach2022high} to synthesis a dataset for 3D domain adaptation. However, it often faces issues such as cumbersome data processing and severe bias including pose and identity. To address these issues, we propose an efficient framework for generalized and consistent 3D domain adaptation, as shown in \cref{fig:method}.

To enable generalized adaptation, drawing inspiration from the impressive performance of Score Distillation Sampling (SDS)~\cite{poole2022dreamfusion} in text-guided 3D generation, we propose multi-modal depth-aware SDS loss (DSDS) for 3D generative domain adaptation. Technically, given a text prompt $T$ or one-shot image reference $I$, we encode it as the condition $c$ in DSDS loss using frozen CLIP's text encoder or image encoder with pre-trained IP-Adapter~\cite{ye2023ip}.

To tackle the inherent overfitting problem in SDS~\cite{kim2022datid}, we introduce depth-aware conditioning from the source generator branch with the same noise. Specifically, we leverage the depth-map $d$ from the volume rendering module of the source generator with pre-trained ControlNet~\cite{zhang2023adding} to preserve the depth information of the source image and mitigate overfitting. Besides, to focus more on foreground adaptation and preserve the background information, we further use the foreground mask $M$ from the depth map of the target generator branch to restrict the adaptation. Formally, DSDS optimizes $\theta$ of $G_{\mathcal{T}}$ by:
\begin{equation} \label{eq:dsds}
\nabla_{\theta} \mathcal{L}_{\text{DSDS}}(\phi, G_{\mathcal{T}}) = \mathbb{E}_{t, \epsilon} \left[w_t \left[ M \cdot \left(\epsilon_{\phi}(z_t; c, d, t) - \epsilon \right) \right] \frac{\partial x}{\partial \theta} \right],
\end{equation}
where $c$ is the condition embedding of the text prompt or image reference, $d$ is the depth map from the source generator, and $\epsilon_{\phi}$ is the denoising UNet of the diffusion model.

\begin{table}[t]
\small
\begin{center}
\setlength{\tabcolsep}{2.2pt}
\renewcommand\arraystretch{1}
\begin{tabular}{ccccccccc}
\toprule
\multicolumn{1}{c}{\multirow{2}{*}{Method}}& \multicolumn{4}{c}{\texttt{Text-Guided}} & \multicolumn{4}{c}{\texttt{Image-Guided}} 
\\
\cmidrule(lr){2-5} \cmidrule(lr){6-9}
    \multicolumn{1}{c}{}
    & {\scriptsize{Pose} ($\downarrow$)} 
    & {\scriptsize{SCS} ($\uparrow$)} 
    & {\scriptsize{CS-T} ($\uparrow$)}
    & {\scriptsize{IS} ($\uparrow$)} 
    & {\scriptsize{Pose} ($\downarrow$)} 
    & {\scriptsize{SCS} ($\uparrow$)} 
    & {\scriptsize{CS-I} ($\uparrow$)}
    & {\scriptsize{IS} ($\uparrow$)} 
    \\ \midrule 
NADA*
&4.403 & 0.481
&28.77 & 1.52
&2.758 &0.565
&84.9 &1.73
 \\ 
DiFa* 
&-	&-
&- &-
&3.305 &0.584
&81.67 &1.70
 \\ 
Fusion* 
&6.536 &0.688
&29.17 &1.44
&-	&-
&-	&-
 \\ 
DATID
&6.102 &0.508
&29 &1.56
&7.815	&0.466
&89.23	&1.74
 \\ 
Ours   
&\textbf{0.768} &\textbf{0.838}
&\textbf{29.5} &\textbf{1.63}
&\textbf{0.926} &\textbf{0.666}
&\textbf{89.24} &\textbf{1.80}
 \\ 
 \bottomrule

\end{tabular}
\end{center}
\vspace{-0.4cm}
\caption{Quantitative comparison with existing text-guided and image-guided domain adaptation methods. Our GCA-3D surpasses baseline methods in terms of pose accuracy, spatial consistency, reference alignment, and diversity.}
\label{tab:eval}
\vspace{-0.6cm}
\end{table}

\begin{figure*}[ht]
\centering
\includegraphics[width=\linewidth]{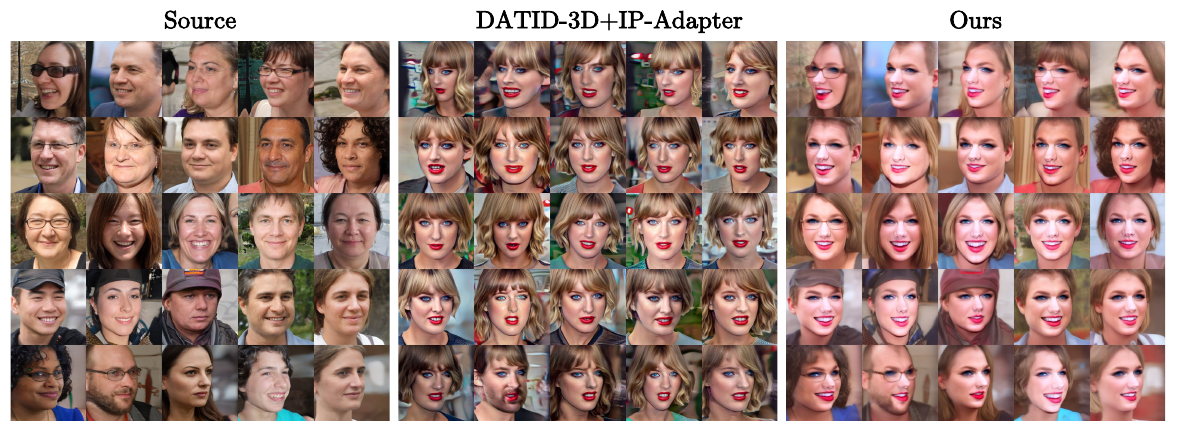}
\vspace{-0.6cm}
\caption{Qualitative comparison with image-guided adversarial adaptation method. We extend DATID-3D with IP-Adapter to enable image-guided adaptation, which suffers from poor pose accuracy and diversity. In contrast, our method generate diverse samples with excellent pose accuracy and identity consistency. Here we use the same image reference in \cref{fig:img}.}
\label{fig:datid}
\vspace{-0.4cm}
\end{figure*}

\begin{figure*}[ht]
\centering
\includegraphics[width=\linewidth]{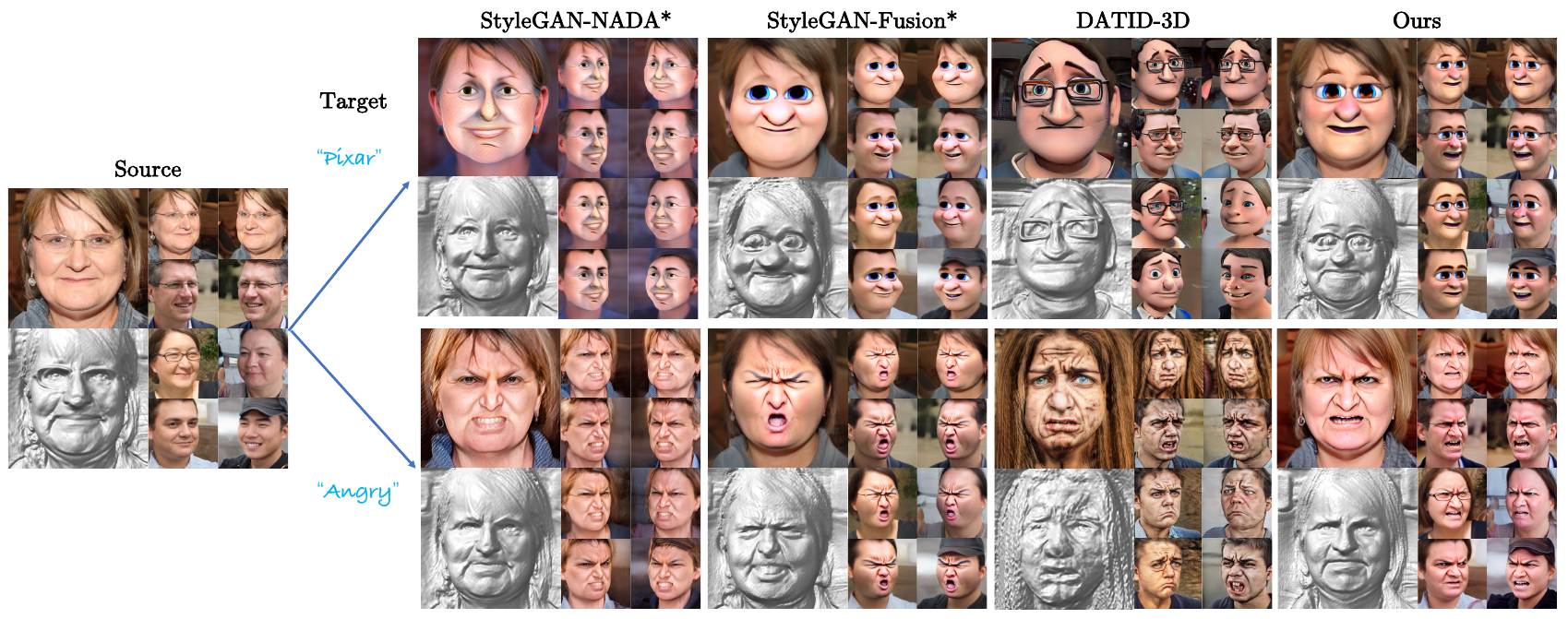}
\vspace{-0.6cm}
\caption{Qualitative comparison with existing text-guided domain adaptation methods. Our GCA-3D yields diverse samples with robust pose accuracy and identity consistency, while other baselines did not.}
\label{fig:text}
\vspace{-0.4cm}
\end{figure*}

\begin{figure*}[ht]
\centering
\includegraphics[width=\linewidth]{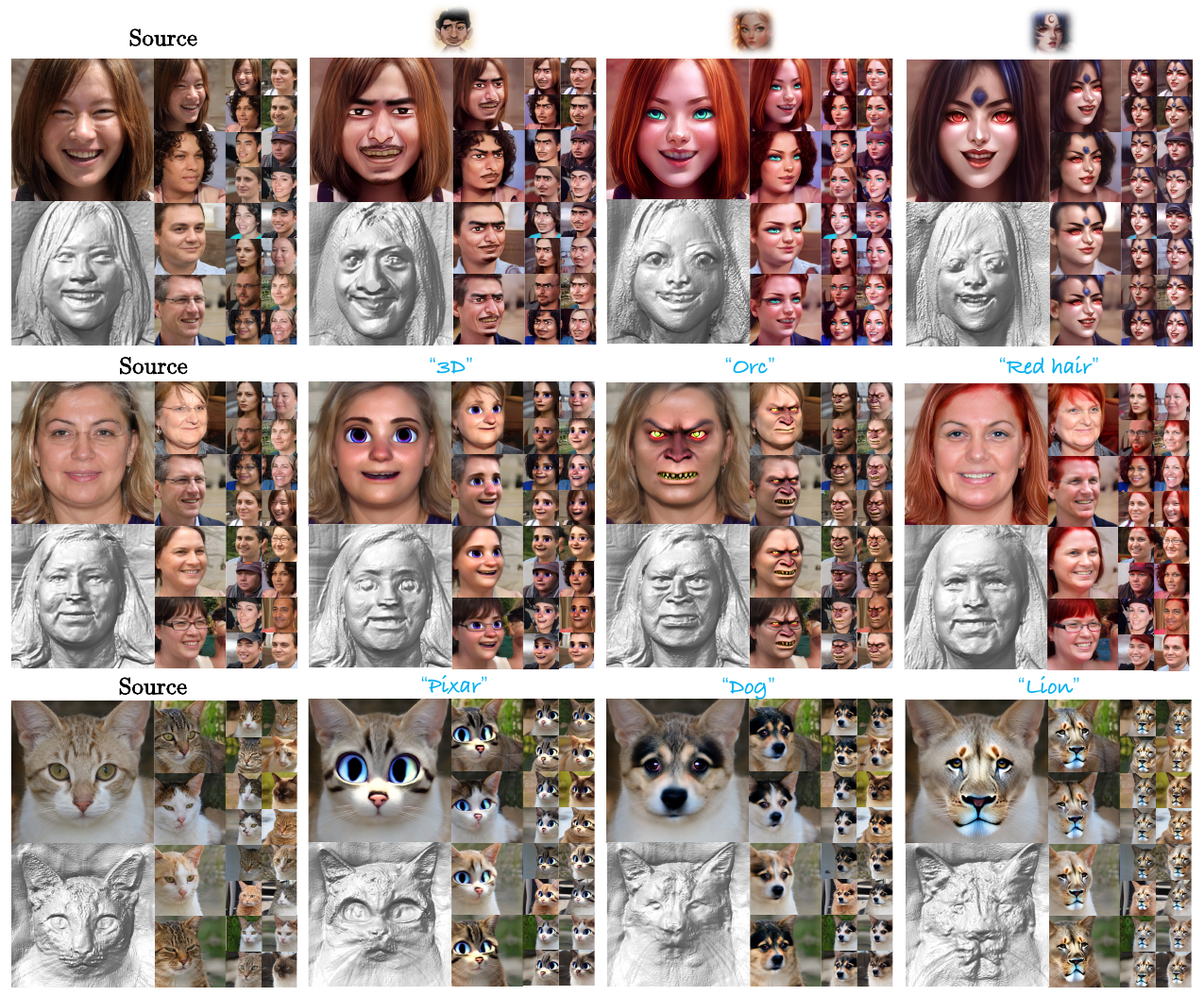}
\vspace{-0.7cm}
\caption{Wide range of generalized adaption results including FFHQ and AFHQ-Cat.}
\label{fig:more}
\vspace{-0.5cm}
\end{figure*}

\begin{figure*}[ht]
\centering
\includegraphics[width=\linewidth]{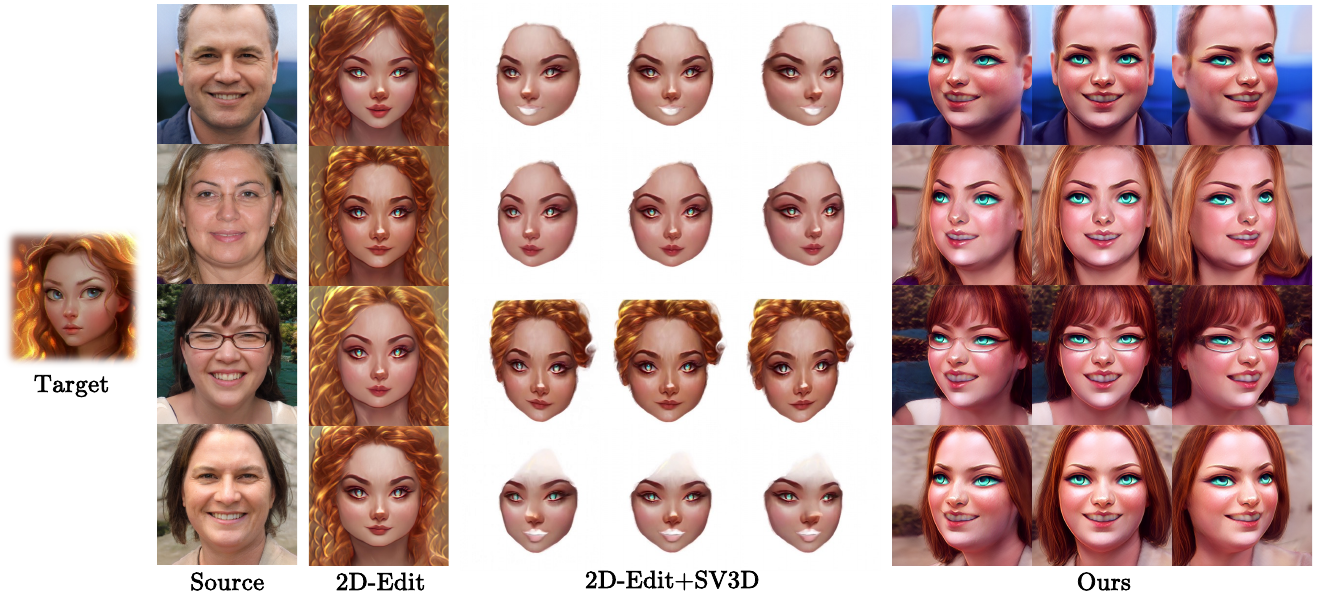}
\vspace{-0.3cm}
\caption{Comparison between GCA-3D and 2D image editing with image-to-3D approach. Specifically, we use StyleCLIP~\citep{patashnik2021styleclip} to perform 2D edits on images and SV3D~\citep{voleti2025sv3d} to convert the edited images to multi-view images.}
\label{fig:edit}
\vspace{-0.4cm}
\end{figure*}

\begin{figure*}[ht]
\centering
\includegraphics[width=\linewidth]{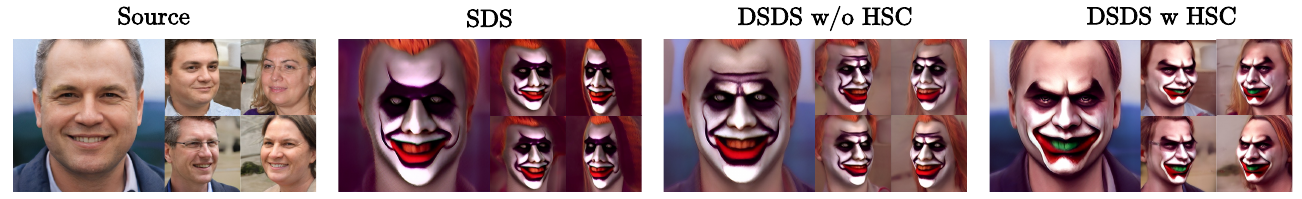}
\vspace{-0.5cm}
\caption{Qualitative ablation study of our method. Here we use the same image reference in \cref{fig:visual}.}
\label{fig:ablation}
\vspace{-0.4cm}
\end{figure*}

\subsection{Hierarchical Spatial Consistency Loss}
Existing 3D domain adaptation methods inevitably introduce pose bias inherent in text-to-image diffusion models. Compared to text-guided adaptation, one-shot image-guided domain adaptation presents greater challenges in two key aspects. (1) It tends to introduce a more significant pose bias during the adaptation process, which stems from the specific 3D pose of the image reference itself. This, in turn, affects the pose accuracy of the adapted generator. (2) Since there is only a single image reference, the adapted model is prone to overfitting certain attributes of the image, such as the closed eyes in \cref{fig:bias}, making it difficult to maintain identity consistency.

To eliminate the bias including pose and identity, we further propose a hierarchical spatial consistency loss. Our objective is to maintain the spatial structure of the generated images between the source generator and the target generator. To this end, we construct the contrastive branch of the source generator that starts from the same noise. By hierarchically aligning the images on the source and target branches on diverse spatial structures, the generator will preserve the pose and identity consistency with the source domain. 

We choose to align the contrastive branches in the embedding space of MViTv2~\cite{fan2021multiscale}. Benefiting from its multi-scale patch-wise tokens with coarse-to-fine features, we could effectively align the hierarchical spatial structure. Thus, our solution is intuitive. We adopt contrastive loss~\cite{oord2018representation}, which aims to reduce the distance between the positive token pairs at the same position and push away the negative token pairs at the other positions by 
\begin{equation}
\begin{gathered}
\mathcal{L}_{\text{HSC}} = -\sum_{i, l}\log\frac{\exp(v_{i, l}^t\cdot v_{i, l}^s)}{\sum_{j}\exp(v_{i, l}^t\cdot v_{j, l}^s)},
\end{gathered}
\end{equation}
where $v_i^t$ and $v_j^s$ are the $i$-th and $j$-th tokens in the $l$-th layer of MViTv2 from $G_\mathcal{T}$ and $G_\mathcal{S}$ respectively. By introducing $\mathcal{L}_{\text{HSC}}$, GCA-3D achieves superior consistency with the source domain including pose and identity.

Thus, the full learning objective is defined as:
\begin{equation}
\begin{gathered}
\mathcal{L}_{\text{overall}} = \mathcal{L}_{\text{DSDS}} + \lambda \mathcal{L}_{\text{HSC}},
\end{gathered}
\end{equation}
where $\lambda$ serves as the hyperparameter that determines the relative importance of HSC loss.
\section{Experiments}
\subsection{Experimental Setting}

\textbf{Baselines.}
Since GCA-3D is generalized to both text-driven and image-driven 3D domain adaptation, we compare it with both types of methods. To the best of our knowledge, our method is the first trial for generalized 3D adaptation tailored for both text-guided and image-guided. Thus, we compare our method with existing image-guided domain adaptation methods (StyleGAN-NADA and StyleGAN-DiFa~\cite{zhang2022towards}) in their 3D extensions. Then we conduct comparisons with text-guided methods such as StyleGAN-NADA~\cite{gal2021stylegan}, StyleGAN-Fusion~\cite{song2022diffusion}, and DATID-3D~\cite{kim2022datid}.

\noindent
\textbf{Evaluation.}
To evaluate our method, we use the EG3D~\cite{chan2022efficient} pretrained on $512^2$ images from the FFHQ dataset~\cite{karras2019style} and AFHQ dataset~\cite{karras2019style} as our source generators, and adapt them to a range of diverse domains. Following EG3D, we measure the pose accuracy estimated from synthesized images by ~\cite{deng2019accurate}. Besides, we use spatial consistency score~\cite{xiao2022few} to compare the spatial consistency with the source domain. We also leverage the CLIP score~\cite{radford2021learning} and IS score~\cite{salimans2016improved} to measure the reference alignment and diversity.  


\subsection{Image-guided 3D Domain Adaptation}
As shown in \cref{fig:visual} and \cref{fig:img}, we conduct the comparison with baseline methods for image-guided domain adaptation. Compared to text-guided adaptation, a single-shot image reference could exacerbate the model overfitting, leading to poorer image quality and diversity in StyleGAN-NADA. StyleGAN-DiFa attempts to improve diversity by incorporating an alignment loss, but the results were still suboptimal. In contrast, our method achieves robust identity consistency with the source domain with hierarchical spatial consistency loss, thereby maintaining excellent diversity while effectively transferring to the target domain.

Additionally, we compare it with DATID-3D with IP-Adapter to validate the superiority over pipeline methods.
As shown in \cref{fig:datid}, DATID-3D exhibits poor pose accuracy and struggles with maintaining identity consistency, leading to reduced diversity in the results. In contrast, our approach significantly outperforms DATID-3D in both pose accuracy and identity consistency. We also conduct quantitative experiments in \cref{tab:eval}, and the results demonstrate that our method significantly outperforms the baselines in terms of pose accuracy, reference alignment, and diversity.

\subsection{Text-guided 3D Domain Adaptation}
As shown in \cref{fig:visual} and \cref{fig:text}, the generator adapted by the StyleGAN-NADA fails to generate high-quality samples, preserving low diversity implicit in the text prompt. Although StyleGAN-Fusion has somewhat improved image quality, the diversity remains suboptimal. DATID-3D maintains better diversity by generating datasets, but it suffers from poor pose accuracy and identity consistency due to the bias in the synthetic dataset. In contrast, our approach performs well across image quality, pose accuracy, and diversity. As shown in \cref{tab:eval}, the quantitative results further corroborate this finding. In addition, we present more visual results including other domains like AFHQ-Cat in \cref{fig:more} to show the generalization of our method.

\begin{table}[t]
\small
\begin{center}
\setlength{\tabcolsep}{4pt}
\renewcommand\arraystretch{1.1}
\begin{tabular}{cccc|ccc}
\toprule
SDS & Depth & Mask & HSC &  Pose ($\downarrow$) & {SCS($\uparrow$)} 
    & {IS ($\uparrow$)} 
    \\ \midrule 
\Checkmark
& & & 
&6.541 &0.691 &1.44 
\\
\Checkmark
&\Checkmark & & 
&4.336 &0.775 &1.56  
\\
\Checkmark
&\Checkmark &\Checkmark & 
&3.928 &0.828 &1.62  
\\
\Checkmark
&\Checkmark &\Checkmark &\Checkmark 
&0.749 &0.842 &1.63  
\\
 \bottomrule

\end{tabular}
\end{center}
\vspace{-0.5cm}
\caption{Quantitative ablation study of our method.}
\label{tab:ablation}
\vspace{-0.4cm}
\end{table}

\subsection{Comparison with Other Approaches}
To demonstrate the superior performance of GCA-3D, we compare it with other alternative approaches. For example, image editing methods, like StyleCLIP~\citep{patashnik2021styleclip}, can effectively edit the images, which can be converted into 3D using Image to 3D methods like SV3D~\citep{voleti2025sv3d}. However, as shown in \cref{fig:edit}, it presents two issues : (1) The inversion in editing inevitably brings information loss, leading to decreased identity consistency as well as degraded diversity, especially for the image condition. (2) While SV3D can lift the edited images to multi-view images, it generates a poor appearance compared with our GCA-3D.

\subsection{Ablation Study}
As shown in \cref{fig:ablation} and \cref{tab:ablation}, we conduct qualitative and quantitative ablation studies of our method. We can observe that DSDS loss effectively mitigates the overfitting problem and preserves the diversity from the source domain. Specifically, depth control improves the spatial consistency and foreground mask mainly maintains the background, which greatly improves SCS and IS score. On the other hand, HSC loss further significantly improves pose accuracy as well as identity consistency, which facilitates more diverse results.

\section{Conclusion \& Limitation}
In this paper, we introduce GCA-3D, a generalized and consistent 3D generative domain adaptation method that eliminates the need for complex data generation pipelines. Unlike previous approaches, GCA-3D employs a multi-modal depth-aware score distillation sampling loss, allowing for efficient non-adversarial adaptation with promising diversity. Additionally, we propose a hierarchical spatial consistency loss to maintain cross-domain consistency, particularly in pose accuracy and identity consistency. We believe our work is an important step towards generalized 3D domain adaptation for the community. 

However, our method also has some limitations. It depends on the capabilities of pre-trained diffusion models and the limitations of performance inherent in these models may also affect the performance of our approach. Nevertheless, we believe that the exploration of generalized 3D domain adaptation is significant for the future work.
{
    \small
    \bibliographystyle{ieeenat_fullname}
    \bibliography{main}
}

\clearpage
\setcounter{page}{1}
\maketitlesupplementary

\begin{figure*}[ht]
\centering
\includegraphics[width=.88\linewidth]{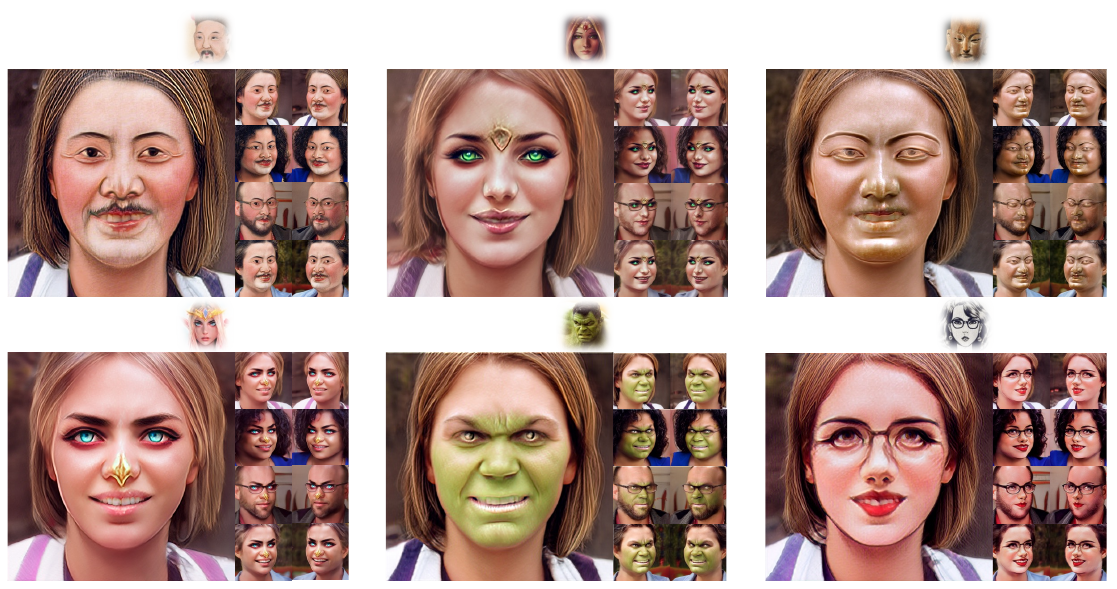}
\vspace{-0.2cm}
\caption{More qualitative results for image-driven domain adaptation.}
\label{fig:more_img}
\vspace{-0.2cm}
\end{figure*}

\begin{figure*}[ht]
\centering
\includegraphics[width=.9\linewidth]{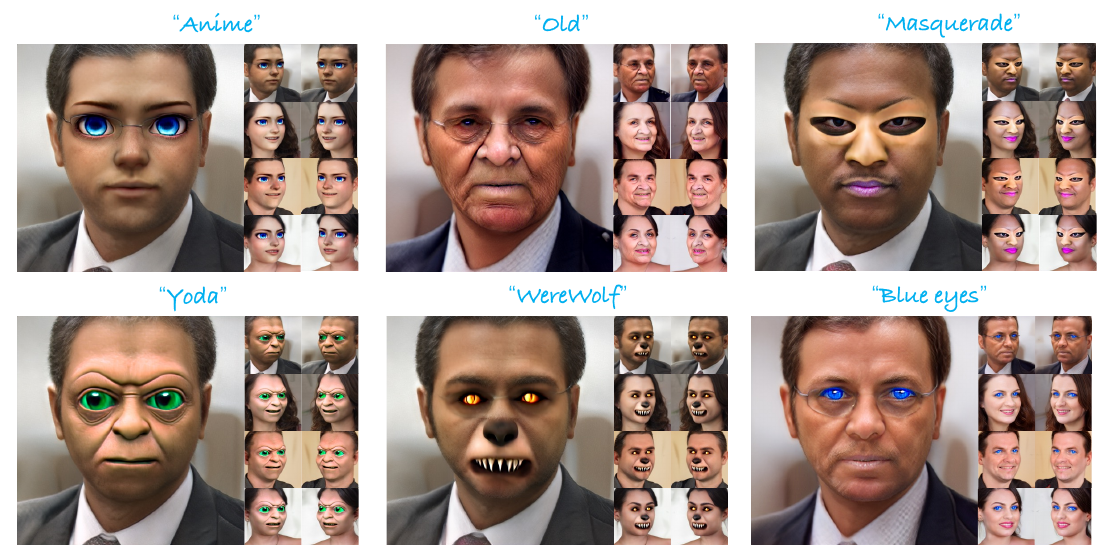}
\caption{More qualitative results for text-driven domain adaptation.}
\label{fig:more_text}
\end{figure*}

\section{Appendix}
\subsection{Additional Results}
\textbf{User Study.}
In the user study, 15 volunteers were asked to evaluate each fine-tuned model on a scale of 1 to 5 across three dimensions: pose accuracy, reference correspondence, image quality, and diversity. We use the EG3D pre-trained in FFHQ and diverse target domains referenced by text or image to generate 1000 images with different seeds for evaluation. Our results, presented in \cref{tab:user}, demonstrate the superior pose accuracy, reference correspondence, diversity, and quality compared to the baselines.

\noindent
\textbf{Videos.}
We include a supplementary video that effectively visualizes and demonstrates how our method, GCA-3D, enables the adapted generator to produce multi-view consistent images with high fidelity and diversity across various text-guided and image-guided target domains.

\begin{table}[t]
\small
\begin{center}
\setlength{\tabcolsep}{2.5pt}
\renewcommand\arraystretch{1}
\begin{tabular}{ccccccccc}
\toprule
Method & Pose & Correspondence & Quality & Diversity
    \\ \midrule 
NADA*
&2.21 & 3.21
&2.03 & 1.89
 \\ 
DiFa* 
&2.33 &3.33
&2.22 &2.10
 \\ 
Fusion* 
&2.83 &3.66
&3.12 &2.23
 \\ 
DATID
&2.01 &3.81
&3.33 &2.89
 \\ 
Ours   
&\textbf{3.58} &\textbf{3.90}
&\textbf{3.50} &\textbf{3.40}
 \\ 
 \bottomrule

\end{tabular}
\end{center}
\vspace{-0.2cm}
\caption{User Study.}
\label{tab:user}
\vspace{-0.2cm}
\end{table}

\begin{table}[t]
\small
\begin{center}
\setlength{\tabcolsep}{2.5pt}
\renewcommand\arraystretch{1}
\begin{tabular}{ccccc}
\toprule
$\lambda$ & 0 & 1 & 3& 5
    \\ \midrule 
Pose($\downarrow$) &3.928 & 1.285 &0.749 &0.723
 \\ 
 CLIP($\uparrow$) &29.85 & 29.52 &29.33 &28.88
 \\
 \bottomrule
\end{tabular}
\end{center}
\vspace{-0.2cm}
\caption{Effectiveness of the coefficient $\lambda$ of HSC loss.}
\label{tab:lambda}
\vspace{-0.2cm}
\end{table}

\noindent
\textbf{More Results.}
\cref{fig:more_img} and \cref{fig:more_text} present additional results of image-driven and text-driven 3D domain adaptation using the EG3D generator, pre-trained on FFHQ. Our GCA-3D enables the generation of diverse, high-fidelity, and multi-view consistent images across various text-guided and image-guided domains, extending beyond the original training domains, all without requiring extra images or knowledge of camera distribution.

\subsection{Implementation Details.}
\textbf{Diffusion Models.} We utilize Stable Diffusion as our diffusion model in DSDS loss, which is based on latent diffusion and employs a pre-trained 123M CLIP ViT-L/14 text encoder to incorporate text prompt conditioning. This diffusion model combines an 860M UNet with the text encoder and image encoder like IP-Adapter, making it lightweight and capable of performing text-conditioned or image-conditioned image synthesis on a GPU with just 10GB of VRAM. We work with Stable Diffusion v1.5. Our code will be made public.

\begin{table*}[!ht]
\begin{adjustbox}{width=0.9\linewidth}
\begin{tabular}{ccc}
\toprule
\multicolumn{1}{l}{Source data type} & Concise prompt & Full text prompt \\
\midrule
\multirow{12}{*}{FFHQ} 
 & Pixar & \textit{"a 3D render of a face in Pixar style"} \\
 & Orc & \textit{"a FHD photo of a face of an orc in fantasy movie"} \\
 & Masquerade & \textit{"a FHD photo of a face of a person in masquerade"} \\
 & Super Mario & \textit{"a 3D render of a face of Super Mario"} \\
 & Yoda & \textit{"a FHD photo of a face of Yoda in Star Wars"} \\
 & Werewolf & \textit{"a FHD photo of a face of Werewolf"} \\
 & Old & \textit{"a FHD photo of a old face"} \\
& Angry & \textit{"a FHD photo of a angry face"} \\
& Red hair & \textit{"a FHD photo of a face with red hair"} \\
& Blue eyes & \textit{"a FHD photo of a face with blue eyes"} \\
& Anime & \textit{"A very beautiful anime girl, full body, long braided curly silver hair, sky blue eyes, full round face, short smile, "} \\
& 3D & \textit{"3d human face, cute big circular reflective eyes, Pixar render"} \\
 \midrule
\multirow{4}{*}{AFHQ Cats}
 & Lion & \textit{"a photo of a lion"} \\
 & Dog & \textit{"a photo of a dog"} \\
 & Pixar & \textit{"3d cat, closeup cute and adorable, cute big circular reflective eyes, Pixar render"} \\
 & Fox & \textit{"a photo of a fox"} \\
\bottomrule
\end{tabular}
\end{adjustbox}
\centering
\caption{List of full text prompts corresponding to each text prompt.}
\label{tab_text_prompt}%
\end{table*}

\noindent
\textbf{Fine-tuning Details.} We fine-tune the 3D generative models using a batch size of 1, training them for 20000 iterations, which are only one-tenth of DATID-3D~\citep{kim2022datid}. The learning rate is set at 0.002. We use MViTv2-L in our HSC loss, and the coefficient $\lambda$ is set to 3. 

\noindent
\textbf{Text Prompts.} In both the main paper and supplementary material, we refer to each text prompt using a concise label. The complete text prompts associated with these concise labels are detailed in \cref{tab_text_prompt}.

\subsection{Trade-off of the Coefficient in HSC Loss}
The coefficient $\lambda$ of HSC loss is the important hyperparameters to determine the degree of spatial consistency with the source domain. We observe that there is a trade-off between the alignment of the target domain and consistency with source domain, which is influenced by $\lambda$. As shown in \cref{tab:lambda}, larger $\lambda$ results in a lower CLIP score, but a higher pose consistency. Empirically, we set $\lambda$ to 3 in all the experiments.

\end{document}